\documentclass[11pt]{article}

\usepackage[preprint]{acl}

\usepackage{times}
\usepackage{latexsym}

\usepackage[T1]{fontenc} 
\usepackage[utf8]{inputenc} 
\usepackage{microtype} 
\usepackage{inconsolata} 
\usepackage{graphicx} 
\usepackage{amsmath}
\usepackage{amsthm}
\usepackage{booktabs}
\usepackage{algorithm}
\usepackage{algorithmic}
\usepackage{multirow} 
\usepackage{amsmath, amssymb} % 添加这两个宏包
\usepackage{booktabs}
\usepackage{tcolorbox}
\tcbuselibrary{breakable}
\usepackage{tcolorbox}
\tcbuselibrary{skins,breakable}
\usepackage{listings}
\usepackage{float}
\usepackage{enumitem}
\usepackage{subfigure}   % 如果模板老一点
\usepackage{enumitem}        % 控制 enumerate/itemize 格式
\usepackage{geometry}
\title{Explicit Representation Alignment for Multimodal Sentiment Analysis}

\author{
Baode Wang$^{\dagger}$,
Ziming Wang, 
Huacan Wang, 
Ronghao Chen, 
Biao Wu \\
AgentAlpha \\
$^{\dagger}$Corresponding author
}

\begin{document}
\maketitle
\begin{abstract}

Multimodal affective analysis aims to understand human sentiment and emotion by jointly modeling heterogeneous modalities such as text and images. However, multimodal models often fail to consistently outperform strong text-only baselines, with performance varying significantly across fusion strategies. In this work, we identify representation misalignment between independently pretrained modality encoders as a key bottleneck for effective multimodal learning, and show through controlled experiments that alignment prior to fusion is often more important than fusion complexity. To address this issue, we propose a unified multimodal affective analysis framework that leverages vision–language models (VLMs) to convert visual content into structured textual descriptions, projecting heterogeneous modalities into a shared linguistic space and enabling interpretable text-centric reasoning. To further improve robustness, we introduce a hybrid learning strategy that combines semantic token selection with a batch-level uniformity regularization objective, encouraging a more dispersed and stable global feature space while mitigating noise introduced by VLM-generated descriptions. Experiments on multiple multimodal sentiment and emotion benchmarks show that our method consistently outperforms strong unimodal and multimodal baselines, achieving state-of-the-art performance. Our analysis further highlights the critical role of representation alignment in multimodal affective learning.

\end{abstract}

\begin{figure}[t!]
    \centering
    \includegraphics[width=0.24\textwidth, angle=270]{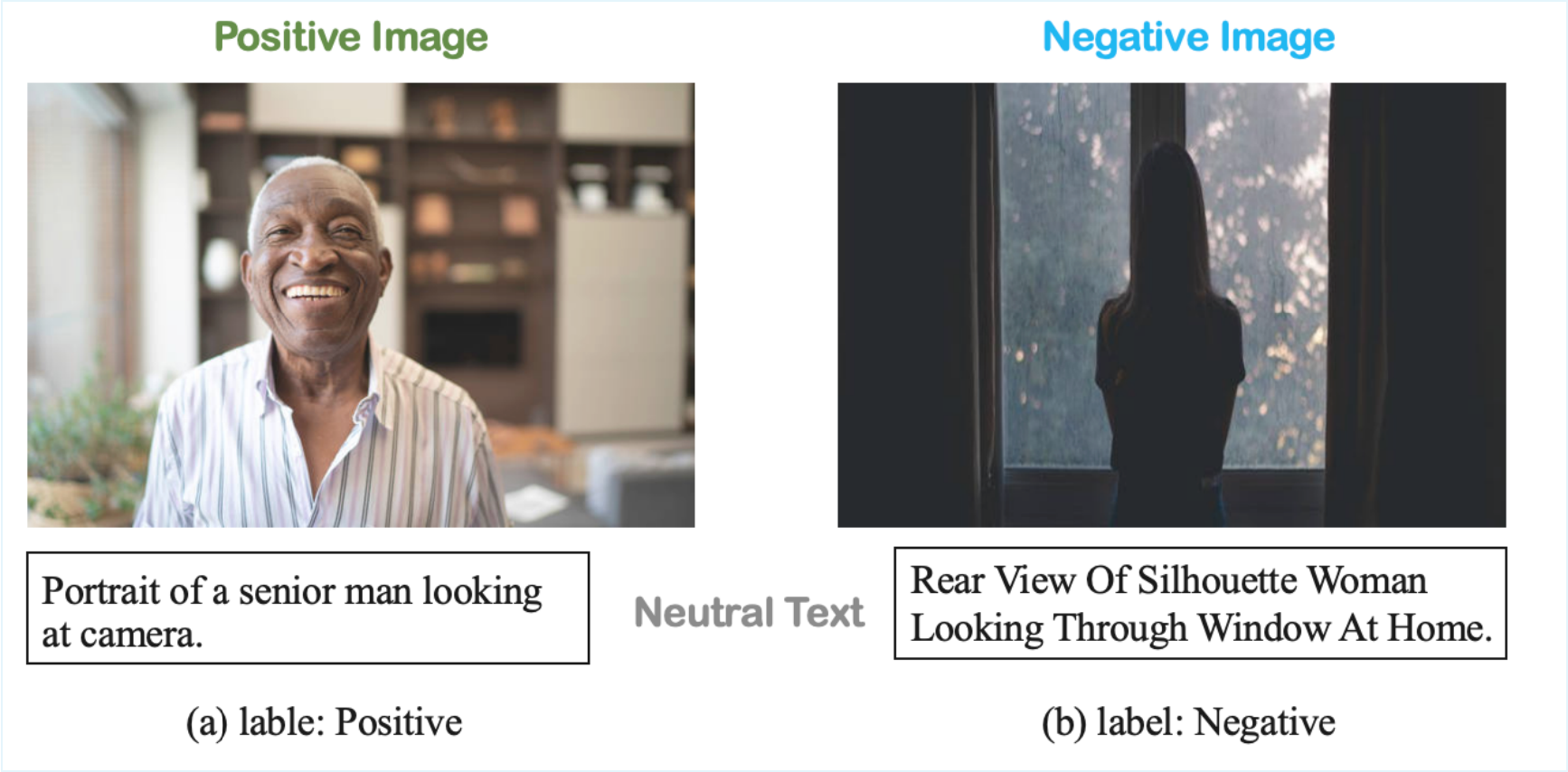}
    \vspace{-3mm}
    \caption{Illustrative examples of multimodal affective classification. The model predicts sentiment polarity by jointly considering the image and its associated text: a clearly smiling face indicates positive affect (left), whereas a darker, implicit scene context suggests negative emotion (right).}
    \label{fig:multimodal_affective_examples}
    \vspace{-3mm}
\end{figure}

\section{Introduction}

Multimodal sentiment analysis and emotion recognition aim to understand human affect by jointly modeling heterogeneous modalities such as text, speech, images, and videos~\citep{xu2018co,jiang2020fusion,rahman2020integrating,yang2020cm,wu2024mmclip,dai2026iwebgenbench,dai2026papervoyager,chen2025paper2web,xu2026idea2story,qian2026story2proposal,wang2025infinityparser}. Compared with text-only analysis, multimodal approaches better reflect real-world communication, where emotions are often conveyed through the interaction of linguistic content, visual context, facial expressions, and vocal cues. Therefore, the same utterance may express different affective meanings under different multimodal contexts, making multimodal affective analysis increasingly important for dialogue systems, user modeling, and emotion-aware human–computer interaction~\citep{zhang2021cross,wu2023mfir,chen2023causal,lu2025multimodal,wu2024mobileagents,dai2026uipress,dai2026papervoyager,yan2025automotiveenv}.

Recent studies have made substantial progress in this area. Early methods mainly relied on modality-specific encoders and fusion architectures to integrate heterogeneous representations~\citep{portner2020desire,robinson1983emotion,xu2019multi,castro2019towards}. More recent work has shifted toward pretrained VLMs and richer cross-modal interaction mechanisms, such as graph-based reasoning, incongruity-aware fusion, and OCR-enhanced multi-view learning. These advances have significantly improved the modeling of cross-modal affective cues and highlight multimodal fusion as a promising direction in affective computing~\citep{guo2025deepseek,liu2024deepseek,team2023gemini,anthropic2024claude4,wang2025infinityparser}.

Despite the rapid progress in multimodal affective analysis, an intriguing and underexplored issue remains: multimodal models do not consistently outperform strong text-only baselines. In practice, while some multimodal approaches achieve clear improvements over unimodal text models, others provide only marginal gains, sometimes less than one percentage point. This observation is counterintuitive, as multimodal inputs are expected to offer richer emotional cues. Notably, such performance inconsistency persists even when sophisticated fusion architectures are employed, suggesting that fusion complexity alone is insufficient to guarantee effective multimodal learning~\citep{team2023gemini,team2025kimi,li2025xiaomi}.

A closer examination of prior work reveals that a key factor lies in the semantic alignment between modality representations before fusion. When text and visual features are extracted independently using modality-specific pretrained encoders (e.g., BERT~\citep{li2019visualbert} for text and ResNet~\citep{he2016deep}  or ViT~\citep{dosovitskiy2020image} for images), their representations often reside in mismatched embedding spaces, making cross-modal interaction difficult to optimize and causing models to rely primarily on textual signals. In contrast, vision-language models~\citep{li2024llava,wu2026vision} pretrained with explicit cross-modal objectives, such as CLIP~\citep{radford2021learning}, map images and text into a shared representation space, substantially easing multimodal fusion. In particular, when text and image encoders are pretrained independently with different architectures and objectives, their output embeddings tend to exhibit severe misalignment~\citep{qwen25vl,hui2024qwen2,yang2025qwen3}. Such misalignment manifests in multiple aspects, including inconsistent feature scales, mismatched distribution statistics, incompatible similarity geometries, and even implicit rotations or rescaling between semantic spaces. These discrepancies make cross-modal interaction difficult to optimize and can undermine the effectiveness of downstream fusion models.

Building upon these observations, we propose a unified multimodal affective analysis framework that explicitly addresses the representation alignment bottleneck prior to fusion. Our approach leverages VLMs to transform visual content into structured textual descriptions, thereby projecting heterogeneous modalities into a shared linguistic space and converting cross-modal interaction into interpretable text–text reasoning. To further enhance robustness and discriminability, we introduce a hybrid learning strategy that combines semantic token selection with a batch-level uniformity regularization objective, enabling the model to focus on affect-relevant cues while encouraging a more dispersed and stable global representation space against noise. Extensive experiments on multiple multimodal sentiment and emotion benchmarks demonstrate that our method consistently outperforms strong unimodal and multimodal baselines, establishing new state-of-the-art results under comparable settings. These results validate both the importance of explicit representation alignment and the effectiveness of our framework for robust and interpretable multimodal affective understanding.

In summary, the main contributions of this work are as follows:
\begin{itemize}[leftmargin=*, align=left]
  \item We propose a unified multimodal affective analysis framework that explicitly addresses the representation alignment bottleneck prior to fusion, by leveraging vision--language models to convert visual content into structured textual descriptions and enabling text-centric multimodal reasoning.
  \item Extensive experiments on multiple multimodal sentiment and emotion benchmarks demonstrate that the proposed method consistently outperforms strong unimodal and multimodal baselines, achieving state-of-the-art performance under comparable settings.
  \item This work provides a systematic analysis and visualization of the role of representation alignment in multimodal learning, offering empirical evidence that feature misalignment can lead to \textbf{modality bias} and consequently degrade model performance.
\end{itemize}

\section{Related Works}

\subsection{ Multimodal Affective Analysis }

Multimodal sentiment analysis (MSA) aims to bridge the semantic gap across heterogeneous modalities such as text and vision. Early work mainly improved cross-modal interaction during fusion, including the Co-memory Network~\citep{xu2018co} and Fusion-Extraction Network~\citep{jiang2020fusion}. With pre-trained models, Transformer-based methods became dominant: CM-BERT~\citep{yang2020cm} and MAG~\citep{rahman2020integrating} inject visual features into BERT to enhance cross-modal semantic interaction. Subsequent studies further explored fine-grained region--token alignment via multi-level attention, such as ITIN~\citep{zhu2022multimodal} and DMLANet~\citep{yadav2023deep}. Beyond semantic consistency modeling in MSA, multimodal sarcasm detection (MSD) and multimodal fake news detection (MFND) additionally require modeling cross-modal \emph{incongruity}. For MSD, recent approaches move from feature concatenation to explicit contradiction reasoning and debiasing, e.g., Cross-modal Graph~\citep{liang2022multi}, DynRT~\citep{tian2023dynamic}, DIP~\citep{wen2023dip}, and DMSD-CL~\citep{jia2024debiasing}. MFND similarly emphasizes inconsistency-aware reasoning, evolving from joint representation learning (MVAE~\citep{khattar2019mvae}) to debiasing and causal/contrastive strategies such as CAFE~\citep{chen2022cross}, multimodal contrastive learning~\citep{zhang2021cross}, MFIR~\citep{wu2023mfir}, and counterfactual causal intervention~\citep{chen2023causal}. As summarized in~\citep{lu2025multimodal}, despite different applications, MSA, MSD, and MFND share a common need to jointly model semantics and affect, motivating unified multimodal perception frameworks (e.g., MDPF) for integrating these tasks.

\subsection{From Emotion to Desire }

High-quality datasets are essential for multimodal affective computing. Over the past decade, several influential benchmarks have been released, including MOSI/MOSEI~\citep{jia2022beyond,zadeh2018multimodal} for sentiment analysis, MELD/IEMOCAP~\citep{poria2019meld,busso2008iemocap} for emotion recognition in multi-party conversations, and task-specific datasets such as Multi-ZOL and MUStARD~\citep{xu2019multi,castro2019towards} for fine-grained sentiment and multimodal sarcasm detection. However, most existing resources emphasize surface-level sentiment polarity and basic emotions, while largely neglecting the deeper cognitive motivation behind affective expressions---\emph{desire}~\citep{jia2022beyond,robinson1983emotion}. Psychological theories view desire as a primitive driver of emotional responses~\citep{portner2020desire,robinson1983emotion}, and Stevens’ sixteen basic desires~\citep{reiss2004multifaceted} highlight intrinsic needs (e.g., curiosity and social contact) that shape cognition and behavior. Despite investigations in psychology and neuroscience (e.g., fMRI)~\citep{cacioppo2012common,hoppe2015recognition}, computational modeling of desire in NLP remains underexplored due to limited datasets~\citep{lim2012desire}.  To address this gap, Jia et al.~\citep{jia2022beyond} introduced MSED, the first multimodal benchmark jointly annotating sentiment, emotion, and desire. MSED contains 9,190 image--text pairs and defines six desire categories (e.g., Romance, Vengeance, and Curiosity). Their analysis further suggests hierarchical correlations among desire, emotion, and sentiment, where desire often implicitly drives emotion and subsequently influences sentiment perception. Therefore, incorporating desire understanding can complement sentiment/emotion modeling and move multimodal affective learning toward more human-like cognition.

\section{Method}
\label{sec:method}

% \subsection{Task Definition}

A key challenge in multimodal affective analysis lies in the heterogeneity of visual and textual representations. Existing multimodal fusion models often struggle to establish stable cross-modal correspondences, causing the training process to degenerate into text-dominated learning. We attribute this limitation to the semantic gap between visual features and linguistic representations, which complicates effective fusion.

To address this issue, we propose a unified multimodal affective framework (UMAF) consisting of three components: (1) a unified fusion backbone that enables controlled comparisons across different encoder combinations; (2) VLM-driven explicit alignment, which converts visual content into structured textual descriptions; and (3) noise-robust feature enhancement, which mitigates VLM-induced noise through Top-K token selection and a uniformity-based regularization objective.

% Given a multimodal sample $(I, T)$, where $I$ denotes an image (or a set of video frames) and $T$ denotes the associated text (e.g., social media posts or conversational utterances), the goal is to predict an affective label $y \in \mathcal{Y}$, such as sentiment polarity or sarcasm category.

% An overview of the framework is illustrated in Figure~\ref{fig:framework}. The image $I$ is processed by a vision--language model (VLM) to generate a structured semantic description $T_{\text{vlm}}$, which serves as a semantic proxy for visual content. The original text $T$ and $T_{\text{vlm}}$ are concatenated and encoded by a Transformer-based text backbone. Finally, we optimize the model with a supervised classification loss and an auxiliary contrastive regularizer to improve robustness and discriminability.

%------------------------------------------------

\subsection{Multimodal Representation Extraction}

As illustrated in Figure~\ref{fig:multimodal_prompt}, we design a structured multimodal reasoning prompt to guide systematic joint analysis of visual and textual inputs for affective understanding. Rather than relying on free-form generation or implicit alignment, the prompt explicitly defines the reasoning goals and procedure, encouraging the model to infer emotional expression, implicit intent, and potential irony or sarcasm via cross-modal integration instead of text-dominated or shallow unimodal reasoning. Concretely, the process is organized into four stages: (i) OCR understanding, which extracts and interprets in-image text with its emotional/semantic implications; (ii) visual scene analysis, which examines cues such as tone, atmosphere, attitude, and symbolic elements; (iii) cross-modal integration, which jointly assesses the consistency, complementarity, or incongruity among external text, OCR text, and visual content; and (iv) high-level reasoning, which incorporates background knowledge and commonsense to infer communicative intent beyond surface sentiment. The prompt further constrains the output into predefined fields (e.g., OCR interpretation, external knowledge, and image–text relations), improving completeness, consistency, and traceability by making intermediate results explicit and verifiable, thereby providing a robust and interpretable framework for multimodal affective inference in scenarios with implicit intent and cross-modal ambiguity.

\begin{figure*}[t!]
    \centering
    \includegraphics[width=0.95\textwidth]{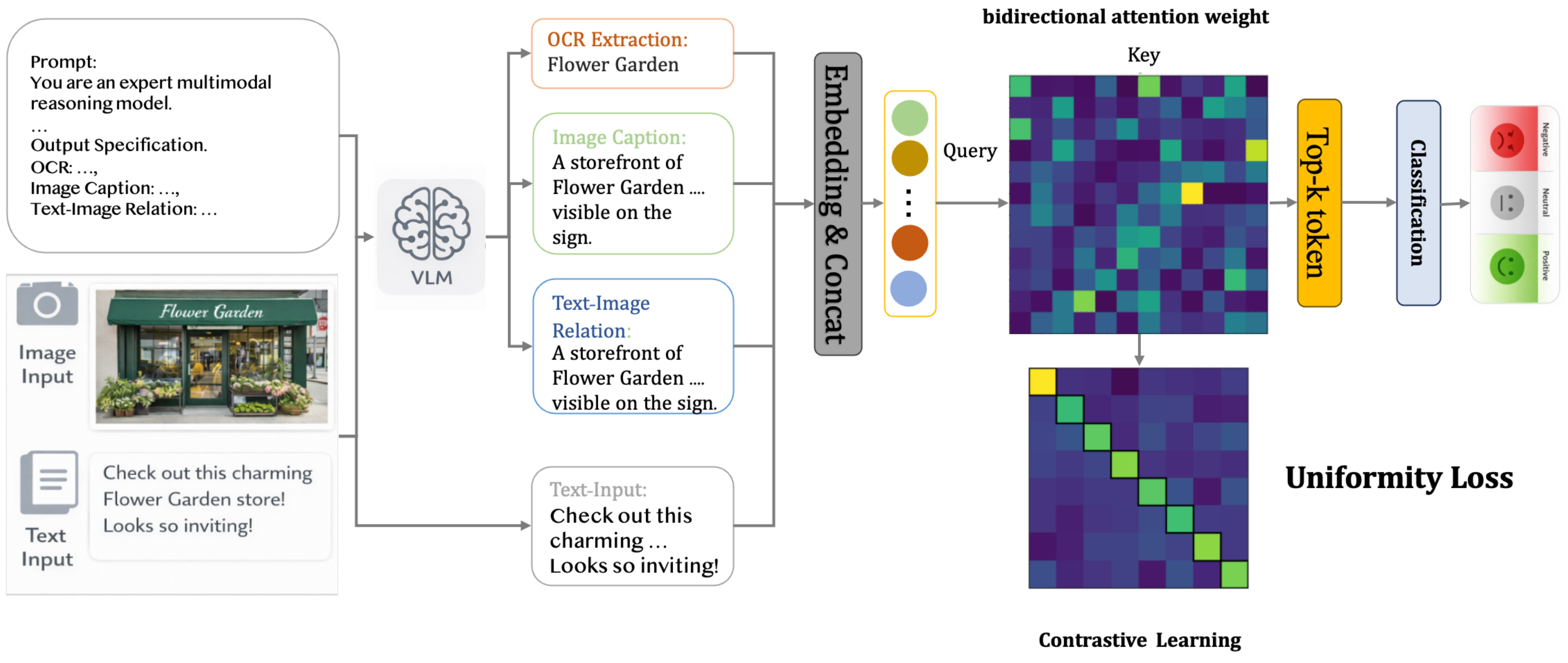}
    \vspace{-3mm}
    \caption{Architecture of the proposed text-centric multimodal framework. Visual content is translated into textual proxies by a VLM and fused with text via RoBERTa and lightweight Transformer layers. Top-K semantic selection supports affective prediction, while batch-level uniformity regularization enhances representation robustness.}
    \label{fig:multimodal_affective_examples}
    \vspace{-3mm}
\end{figure*}

\subsection{Cross-Modal Fusion}
\label{sec:fusion}

Conventional multimodal fusion methods typically rely on cross-attention mechanisms that treat visual representations as keys and values conditioned on textual queries. While effective, such designs often suffer from representation heterogeneity and unstable alignment, especially when visual features are noisy or weakly correlated with linguistic semantics. To address this issue, we adopt a \emph{text-centric fusion strategy} that unifies visual and textual information within a single linguistic representation space.

Specifically, visual content is first distilled into structured textual descriptions via a vision--language model (VLM), yielding a visual proxy representation \(T_{\text{vlm}}\). By converting visual information into language-aligned tokens, both modalities can be jointly modeled using a single pretrained language encoder, avoiding explicit cross-modal attention while reducing the semantic gap between modalities.

The original textual input \(T\) and the VLM-generated descriptions \(T_{\text{vlm}}\) are concatenated at the token level to form a unified input sequence, which is then encoded by a pretrained RoBERTa backbone:
\vspace{-3mm}
\begin{equation}
    \mathbf{H}_{\text{base}} = \text{RoBERTa}\bigl([T; T_{\text{vlm}}]\bigr) \in \mathbb{R}^{L \times d_h},
\end{equation}
where \([T; T_{\text{vlm}}]\) denotes token concatenation, \(L\) is the total sequence length, and \(d_h\) is the hidden dimension of RoBERTa (\(d_h = 768\) for RoBERTa-base).

Although \(\mathbf{H}_{\text{base}}\) implicitly incorporates visual semantics through \(T_{\text{vlm}}\), we observe that interactions between the original text and its visual proxy remain largely implicit at this stage. Such implicit fusion may be insufficient for capturing subtle affective cues, including emotional incongruity, irony, or visually grounded sentiment shifts. To enable more explicit and fine-grained cross-modal interaction, we introduce two lightweight Transformer layers on top of the RoBERTa representations:

\vspace{-2mm}
\begin{equation}
    \mathbf{Z} = \text{Transformer}_{\text{fusion}}(\mathbf{H}_{\text{base}}) \in \mathbb{R}^{L \times d_h},
\end{equation}
where \(\text{Transformer}_{\text{fusion}}\) preserves the same hidden dimension \(d_h\) but is randomly initialized and trained from scratch. These task-specific fusion layers learn to emphasize informative cross-token dependencies, explicitly modeling alignments as well as discrepancies between \(T\) and \(T_{\text{vlm}}\), which are crucial for accurate affective reasoning.

% \vspace{2mm}
% \noindent\textbf{Noise-Robust Feature Enhancement.}
\paragraph{Top-K Semantic Selection}
% \noindent\textbf{Top-K Semantic Selection}
While VLM-generated descriptions provide valuable visual semantics, they may also introduce redundant or noisy tokens that are irrelevant to the downstream task. To improve discriminability and robustness against such noise, we further adopt a hybrid training strategy that combines \emph{Top-$K$ semantic token selection} with supervised learning.

Not all tokens in the fused sequence contribute equally to affective prediction. Given the fused representation \(\mathbf{Z}\), we estimate the importance of each token using a lightweight scoring network:

\vspace{-3mm}
\begin{equation}
\mathbf{s} = \sigma(\mathbf{W}_s \mathbf{Z} + \mathbf{b}_s),
\end{equation}
where \(\mathbf{W}_s\) and \(\mathbf{b}_s\) are learnable parameters, and \(\sigma(\cdot)\) denotes the sigmoid activation function. The resulting score vector \(\mathbf{s}\) reflects the relevance of each token to the target affective task.

We then select the indices of the top-$K$ tokens with the highest importance scores, denoted by \(\mathcal{I}_{\text{top}}\), and aggregate their representations via mean pooling:

\vspace{-5mm}
\begin{equation}
\mathbf{h}_{\text{top}} = \frac{1}{K} \sum_{i \in \mathcal{I}_{\text{top}}} \mathbf{Z}_i.
\end{equation}

The aggregated representation \(\mathbf{h}_{\text{top}}\) captures the most informative semantic regions across both the original text and the visual proxy tokens. This representation is subsequently fed into a task-specific classifier and optimized using the standard cross-entropy loss \(\mathcal{L}_{\text{ce}}\). By explicitly supervising the most salient tokens, the model is encouraged to focus on affect-relevant cues while suppressing noisy or irrelevant information introduced during VLM generation.

%------------------------------------------------

\subsection{Batch-Level Uniformity Regularization}

To further regularize the global feature space and enhance representation discriminability, we introduce a batch-level uniformity regularization objective at the sequence level. Unlike contrastive learning methods that explicitly construct positive and negative pairs, our objective does not rely on instance matching or augmentation consistency. Instead, it directly encourages the global representations of different samples within a mini-batch to be more uniformly dispersed in the latent space. In this way, the model is less likely to produce overly concentrated representations, which helps mitigate the negative impact of noisy or redundant information introduced by VLM-generated descriptions.

Specifically, given a mini-batch of \(B\) samples, let \(\bar{\mathbf{z}}_i \in \mathbb{R}^d\) denote the global sequence representation of the \(i\)-th sample, obtained by applying global pooling over the fused token representations \(\mathbf{Z}\). This representation captures the holistic semantics of the corresponding multimodal input. To encourage a more dispersed feature distribution, we compute pairwise similarities among samples in the same mini-batch and penalize cases where different instances are mapped too closely in the representation space.

Based on this intuition, we define the uniformity objective as:

\vspace{-3mm}
\begin{equation}
\small
\mathcal{L}_{\text{uni}}=
\frac{1}{B}\sum_{i=1}^{B}
\log\left(
\sum_{j=1,j\neq i}^{B}
\exp\big(\operatorname{sim}(\mathbf{z}_i,\mathbf{z}_j)/\tau\big)
\right),
\end{equation}

where:
\begin{itemize}
    \item $\mathbf{z}_i$: the feature representation of the $i$-th instance in a mini-batch of size $B$;
    \item $\operatorname{sim}(\mathbf{u}, \mathbf{v})$: the cosine similarity between two instances;
    \item $\tau$: a temperature scaling factor controlling the sensitivity to highly similar sample pairs;
    \item $\mathcal{L}_{\text{uni}}$: the uniformity regularization term, which encourages different instances to stay apart and thus promotes a better distributed feature space.
\end{itemize}

By minimizing this objective, the model is encouraged to reduce excessive similarity among different samples in the batch, thereby preventing feature collapse and improving inter-instance separability. Importantly, this objective does not directly supervise affective classification; instead, it acts as a geometric regularizer on the global representation space. Through backpropagation, all token representations implicitly participate in this regularization process, including noisy or affect-irrelevant components. Meanwhile, the supervised affective classification loss emphasizes task-relevant emotional cues. The combination of these two objectives allows the model to maintain a well-structured global feature space while preserving discriminative information for downstream prediction.

Finally, the overall training objective is defined as a weighted combination of the affective classification loss and the batch-level uniformity regularization term:

\vspace{-3mm}
\begin{equation}
\mathcal{L}_{\text{total}} = \mathcal{L}_{\text{ce}} + \lambda \mathcal{L}_{\text{uni}},
\end{equation}

where \(\lambda\) controls the trade-off between discriminative supervision and feature-space regularization.

\section{ Experiment }
 
\subsection{  Experimental datasets }

In our experiments, we evaluate the proposed method on two widely used multi-modal benchmarks, namely MSED and MVSA, which are designed for affective analysis from paired textual and visual data. These datasets differ in their annotation schemes and task settings, allowing us to comprehensively assess the effectiveness and generalization ability of our approach across multiple multi-modal affective analysis scenarios.

\textbf{MSED} is a multi-modal benchmark designed for comprehensive affective analysis. It contains 9,190 image–text pairs collected from social media platforms such as Twitter, Flickr, and Getty Images. Each sample is manually annotated with six human desire categories (family, romance, vengeance, curiosity, tranquility, and social-contact), three sentiment polarities (positive, neutral, and negative), and six emotion classes (happiness, sad, neutral, disgust, anger, and fear). The dataset supports multi-task learning, including desire detection, sentiment analysis, and emotion recognition, and enables the investigation of the intrinsic correlations among these affective factors in a unified multi-modal setting. MSED is split into training, validation, and test sets with a ratio of 70\% / 10\% / 20\%, providing a standardized evaluation protocol for multi-modal affective modeling.

\textbf{MVSA} is a benchmark dataset designed for multi-modal sentiment analysis, consisting of image–text pairs collected from Twitter with manual sentiment annotations. It comprises two subsets: MVSA-Single, which contains 4,869 image–text pairs after removing samples with inconsistent sentiment labels between image and text, and MVSA-Multiple, which includes 16,779 image–text pairs with multiple annotations to capture diverse human perceptions. MVSA serves as a valuable benchmark for both single-view and multi-view sentiment analysis, and experimental studies on this dataset demonstrate that jointly modeling textual and visual information can effectively improve sentiment analysis performance.

% ~\citep{yadav2023deep}

\subsection{ Baselines }

% We compare the performance of MDPF with a set of strong baseline methods across three tasks, including multimodal sentiment analysis (MSA), multimodal sarcasm detection (MSD), and multimodal fake news detection (MFND), as summarized in Tables 1, 2, and 3. 

For the MVSA task, we adopt several representative multimodal sentiment analysis baselines that jointly leverage image and text information. Specifically, CoMN~\citep{xu2018co} exploits the correlation between image and text modalities to enhance multimodal representation learning, while FENet~\citep{jiang2020fusion} employs an interactive information fusion mechanism to learn vision-specific textual representations and text-specific visual representations. CM-BERT ~\citep{yang2020cm} relies on cross-modal interaction to fine-tune a pre-trained BERT model, and MAG~\citep{rahman2020integrating} allows BERT and XLNet to incorporate multimodal non-linguistic features during fine-tuning. In addition, MVAN~\citep{zhu2022multimodal} utilizes a continuously updated memory network to obtain deep semantic representations of image–text pairs. To model fine-grained cross-modal alignment, ITIN~\citep{zhu2022multimodal} introduces a region–word correspondence module and fuses multimodal features through an adaptive cross-modal gating mechanism, while DMLANet~\citep{yadav2023deep} employs deep multi-level attention to further exploit correlations between image and text modalities.

\begin{table}[t]
\centering
\small
\setlength{\tabcolsep}{4pt}
\resizebox{\linewidth}{!}{
\begin{tabular}{l c c c c c c}
\toprule
\multirow{2}{*}{Method} 
& \multicolumn{3}{c}{Sentiment Analysis} 
& \multicolumn{3}{c}{Emotion Recognition} \\
\cmidrule(lr){2-4} \cmidrule(lr){5-7}
& P & R & F1 & P & R & F1 \\
\midrule
Multi-modal CNN        & 36.52 & 45.65 & 40.60 & 25.01 & 23.38 & 24.17 \\
BiGRU+Att              & 53.29 & 48.07 & 50.54 & 27.49 & 23.12 & 25.11 \\
SVM+BERT               & 67.81 & 66.35 & 67.07 & 62.45 & 50.25 & 55.69 \\
GRU-RoBERTa            & 84.63 & 84.76 & 84.69 & 80.41 & 80.51 & 80.46 \\
EfficientNet           & 73.29 & 74.24 & 73.76 & 59.88 & 60.57 & 60.22 \\
UPB-MTL                & 81.48 & 81.93 & 81.70 & 67.17 & 66.07 & 66.62 \\
MMBT                   & 83.32 & 83.59 & 83.45 & 81.51 & 82.51 & 82.01 \\
A-MTL                  & 83.38 & 84.23 & 83.80 & 80.27 & 80.75 & 80.51 \\
CIM                    & 81.52 & 82.41 & 81.96 & 78.77 & 79.23 & 79.00 \\
Multi-task RNN         & 74.29 & 75.63 & 74.95 & 64.24 & 64.29 & 64.27 \\
M3GAT                  & 84.66 & 85.15 & 84.90 & 82.53 & 81.51 & 82.02 \\
MMTF-DES               & 88.27 & 88.68 & 88.47 & 84.39 & 84.64 & 84.52 \\
\midrule
\textbf{Ours}          & \textbf{89.57} & \textbf{89.34} & \textbf{89.46} & \textbf{87.60} & \textbf{86.58} & \textbf{87.09} \\
\bottomrule
\end{tabular}
}
\caption{Performance comparison on the MSED dataset. 
F1 denotes the harmonic mean of precision (P) and recall (R).}
\label{tab:msed_results}
% \vspace{-3mm}
\end{table}

\subsection{Main Result}

\paragraph{Results on MSED}
Table~1 presents the performance comparison on the MSED dataset for sentiment analysis and emotion recognition. Overall, our method consistently achieves the best results across all evaluation metrics, outperforming all competing baselines on both tasks. For sentiment analysis, although advanced multimodal and interaction-based models such as CIM, M3GAT, and MMTF-DES exhibit strong performance, our approach attains the highest F1-score of \textbf{89.46\%}, indicating more effective multimodal feature integration. Similarly, for emotion recognition, our model achieves the best F1-score of \textbf{87.09\%}, demonstrating superior capability in capturing fine-grained emotional cues. These results validate the effectiveness and robustness of our unified multimodal reasoning framework on the MSED dataset.

\begin{table}[t]
\centering
\renewcommand{\arraystretch}{1.1}
\setlength{\tabcolsep}{4pt}
\resizebox{\linewidth}{!}{
\begin{tabular}{lcc}
\toprule
\textbf{Models} & \multicolumn{2}{c}{\textbf{MVSA-Single}} \\
\cmidrule(lr){2-3}
 & \textbf{Accuracy} & \textbf{F1-score} \\
\midrule
CoMN~\citep{xu2018co}    & 70.5 & 70.0 \\
FENet~\citep{jiang2020fusion}   & 74.2 & 74.0 \\
CM-BERT~\citep{yang2020cm}  & 71.3 & 72.7 \\
MAG~\citep{rahman2020integrating}      & 77.8 & 76.2 \\
MVAN~\citep{zhu2022multimodal}     & 73.1 & 72.3 \\
ITIN~\citep{zhu2022multimodal}   & 75.2 & 75.0 \\
DMLANet~\citep{yadav2023deep} & 79.4 & 79.5 \\
MDPF-CLIP~\citep{LU2025102747} & 80.6 & 80.3 \\
\midrule
\textbf{Ours} & \textbf{82.3} & \textbf{81.8} \\
\bottomrule
\end{tabular}
}
\caption{Performance comparison on the MVSA-Single dataset. 
F1 denotes the harmonic mean of precision (P) and recall (R).}
\label{tab:mvsa-single}
\end{table}

\paragraph{Results on MVSA-Single}
Table~\ref{tab:mvsa-single} summarizes the performance of different multimodal sentiment analysis models on the MVSA-Single dataset. Overall, our method achieves the best results in terms of both accuracy and F1-score, indicating its strong effectiveness in modeling fine-grained image–text sentiment cues. These performance gains demonstrate that stronger cross-modal representation capability and unified multimodal perception play a crucial role in improving sentiment analysis performance on MVSA-Single.

\begin{table}[t]
\centering
\renewcommand{\arraystretch}{1.1}
\setlength{\tabcolsep}{3pt}
\resizebox{0.48\textwidth}{!}{% 调整宽度为文本宽度，高度自动按比例调整
\begin{tabular}{lccccc}
\toprule
\textbf{Encoder} & \textbf{Setting} & \textbf{Acc (\%)} & \textbf{F1 (\%)} & \textbf{Prec (\%)} & \textbf{Rec (\%)} \\
\midrule
BERT             & Text-only        & 83.59   & 83.48  & 83.30  & 83.69  \\
BERT + ResNet    & Multimodal       & 82.95   & 82.88  & 83.00  & 82.78  \\
\midrule
RoBERTa          & Text-only        & 81.68   & 81.58  & 81.33  & 81.06  \\
RoBERTa + ViT    & Multimodal       & 80.60   & 80.18  & 80.56  & 80.77  \\
\midrule
CLIP (Text)      & Text-only        & 84.62   & 84.61  & 84.71  & 84.54  \\
CLIP (Text+Image)& Multimodal       & 85.79   & 85.77  & 85.70  & 85.84  \\
\bottomrule
\end{tabular}%
}
\caption{Performance comparison of different feature extractor combinations under two settings: Text-only, and Multimodal. Accuracy (Acc), F1-score (F1), Precision (Prec), and Recall (Rec) are reported in percentage.}
\label{tab:feature_extractor_metrics_3setting}
% \vspace{-3mm}
\end{table}

\subsection{Ablation Study}

\paragraph{Impact of Modal Alignment}

Table~\ref{tab:feature_extractor_metrics_3setting} reports an ablation study on different feature extractor combinations under text-only, image-only, and multimodal settings. Overall, multimodal configurations consistently outperform unimodal ones, demonstrating the importance of jointly modeling textual and visual information. Text-only models based on BERT, RoBERTa, and CLIP (Text) achieve relatively strong performance, indicating that textual cues play a dominant role in affective understanding. In contrast, image-only models show significantly weaker results, suggesting that visual information alone is insufficient to capture fine-grained affective semantics. When visual features are incorporated into text encoders, performance is further improved, as observed in BERT+ResNet and RoBERTa+ViT, confirming the complementary nature of the two modalities. Among all configurations, the CLIP-based multimodal model achieves the best performance across all evaluation metrics, substantially outperforming both its text-only and image-only variants, which highlights the effectiveness of CLIP’s cross-modal alignment in bridging the semantic gap between image and text. Overall, these results validate that modality complementarity and effective cross-modal representation learning are critical for achieving superior performance.

\paragraph{Impact of Noise-Robust Features}

\begin{table}[t]
\centering
\renewcommand{\arraystretch}{1.1}
\setlength{\tabcolsep}{3pt}
\resizebox{0.45\textwidth}{!}{% 调整宽度为文本宽度，高度自动按比例调整
\begin{tabular}{l l c c c c}
\toprule
Model & Setting  & Prec. (\%) & Rec. (\%)  & F1 (\%)\\
\midrule
UMAF & $K=5$             & 90.75 & 87.99 &   89.35 \\
UMAF & $K=20$            & 87.26  & 87.38  & 87.32   \\
UMAF & w/o NRF           &  89.28 & 87.73 &   88.5 \\
UMAF & Full($K=10$)     &  89.57 &  89.34  & 89.46   \\
\bottomrule
\end{tabular}
}
\caption{Ablation results on the Sentiment Analysis task of the MSED dataset. Accuracy (Acc), F1-score (F1), Precision (Prec), and Recall (Rec) are reported in percentage.}
\label{tab:setting}
% \vspace{-5mm}
\end{table}

\begin{figure*}[t]
    \centering
    \includegraphics[width=1\linewidth]{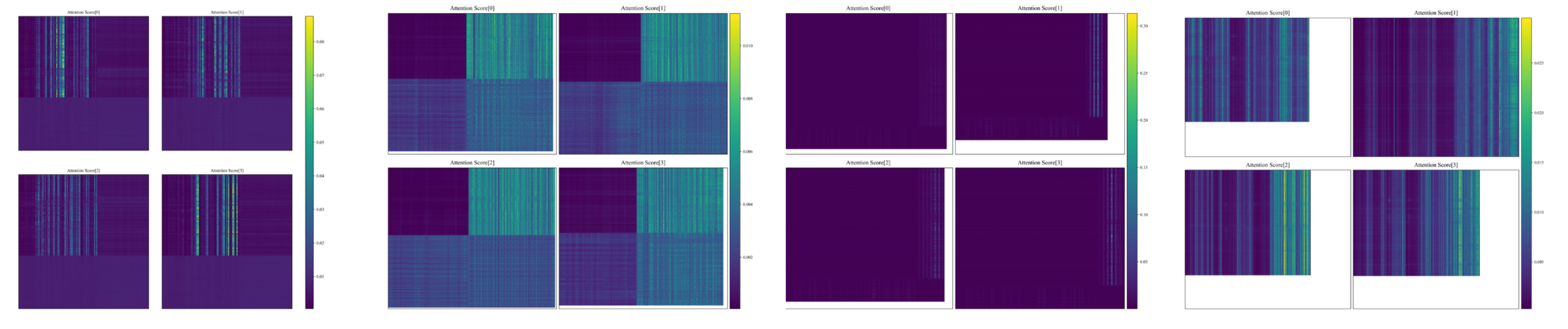}
    \caption{ Visualization of attention score distributions in the attention matrices for different feature extractor combinations. From left to right, the models are BERT+ResNet, RoBERTa+ViT, CLIP, and VLM.}
    \label{fig:visual}
    \vspace{-3mm}
\end{figure*}

As shown in Table~\ref{tab:setting}, the Noise-Robust Features module contributes positively to model performance. Removing this component leads to consistent drops in Precision, Recall, and F1, indicating that it effectively mitigates the noise introduced by VLM-generated content and improves the stability and discriminability of the learned representations. In addition, the results under different Top-$K$ settings suggest that the choice of $K$ is critical. The best performance is achieved when $K=10$, whereas a larger value such as $K=20$ results in noticeable degradation. This suggests that an excessively large $K$ may introduce more redundant semantics that are irrelevant to affective prediction, thereby weakening the effect of salient feature selection. Overall, these findings demonstrate that both noise-robust feature enhancement and an appropriate Top-$K$ selection are important for achieving strong performance in multimodal affective analysis.

% \begin{figure}[t]
%     \centering
%     \includegraphics[width=\linewidth]{figures/pic2.jpeg}
%     \caption{Visualization of attention weights over fused image--text representations. Brighter colors indicate higher attention scores, highlighting salient visual and textual tokens that contribute most to affective reasoning.}
%     \label{fig:attention_vis}
% \end{figure}

% \begin{figure*}[t]
%     \centering
%     \subfigure[]{
%         \includegraphics[width=0.23\linewidth]{figures/pic2.jpeg}
%     }
%     \subfigure[]{
%         \includegraphics[width=0.23\linewidth]{figures/attention_vlm.png}
%     }
%     \subfigure[]{
%         \includegraphics[width=0.23\linewidth]{figures/attention_bert_resnet.png}
%     }
%     \subfigure[]{
%         \includegraphics[width=0.23\linewidth]{figures/attention_roberta_vit.png}
%     }
%     \caption{Visualization of attention weights over fused image--text representations. Brighter colors indicate higher attention scores, highlighting salient visual and textual tokens that contribute most to affective reasoning.}
%     \label{fig:attention_vis}
% \end{figure*}

\section{Further Analysis}

% \subsection{Attention Visualization}
Figure~\ref{fig:visual} presents the attention distributions over the fused image–text representations. Each column corresponds to a token in the unified sequence, and the color intensity reflects its relative attention weight. As can be observed, under the first three configurations, the model exhibits a clearly non-uniform attention pattern across modalities, revealing a pronounced modality bias. Specifically, the attention weights are heavily concentrated on a subset of tokens in the later part of the sequence, which mainly correspond to the visual tokens. This phenomenon suggests that, when the representations of different modalities are not well aligned, the model tends to over-rely on one modality during internal information aggregation, resulting in an imbalanced cross-modal interaction pattern. Such modality-dominated behavior weakens the model’s ability to capture complementary semantics across modalities and consequently limits its overall performance.

In contrast, the last configuration demonstrates a substantially different attention pattern. After visual content is projected into the textual space, it can provide complementary affective cues in a semantically compatible form. Rather than performing naive or uniform fusion, the model allocates attention more selectively across modalities, enabling more effective cross-modal integration. Moreover, the high-attention regions are no longer confined to a single modality but are distributed across both image-derived and text-derived representations. This indicates that the model is capable of focusing on discriminative cross-modal semantic cues, such as emotion-bearing visual entities, contextual descriptions, or semantically aligned textual evidence, while suppressing noisy or affect-irrelevant tokens.

Overall, this visualization provides intuitive evidence that explicit alignment before fusion is critical for multimodal affective analysis. Without proper alignment, the model is prone to modality bias, causing one modality to dominate the decision process and reducing the effectiveness of multimodal interaction. By contrast, explicit alignment encourages the model to establish a more balanced and semantically coherent interaction between modalities, thereby improving both interpretability and predictive performance. These findings further support our central claim that representation alignment is not merely an auxiliary design choice, but a fundamental prerequisite for robust and effective multimodal reasoning.

\section{Conclusion}

In this work, we present a unified multimodal affective analysis framework that explicitly bridges visual and textual semantics through VLM-driven alignment and robust representation learning. By converting visual content into structured textual descriptions via carefully designed prompts, we adopt a text-centric fusion paradigm that unifies multimodal information within a shared linguistic space, effectively alleviating the semantic gap between heterogeneous modalities. To further enhance discriminability and robustness, we introduce a hybrid learning strategy that combines Top-K semantic token selection with batch-level uniformity regularization, enabling the model to focus on affect-relevant cues while promoting a more dispersed and stable global feature space against VLM-induced noise. Extensive experiments on multiple multimodal affective benchmarks demonstrate that our approach consistently outperforms strong baselines across sentiment, emotion, and related tasks, validating the effectiveness of structured visual grounding and uniformity-based regularization. We believe this work provides a principled and extensible direction for integrating large VLMs into multimodal affective understanding, and opens up new opportunities for interpretable and robust multimodal reasoning.

\nocite{Ando2005,andrew2007scalable,rasooli-tetrault-2015}

\section*{Limitations}

Our method still has several limitations. First, it mainly focuses on text–image affective analysis and does not explicitly model other modalities such as speech or video. Second, the framework relies on VLM-generated textual descriptions, whose quality may affect downstream performance when the generated content is noisy or incomplete. Third, although the proposed alignment strategy improves robustness, its effectiveness has only been validated on benchmark datasets, and its generalization to more diverse real-world scenarios remains to be further studied. Finally, the additional VLM-based description generation introduces extra computational cost compared with conventional fusion-based methods.

% \section*{Acknowledgments}

\bibliography{acl_latex}

\appendix

\section{Appendix}

\begin{figure*}[t]
\centering

\begin{tcolorbox}[
  title=Multimodal Reasoning Prompt,
  colframe=black!30,
  colback=black!2,
  fonttitle=\bfseries,
  % breakable,
  width=\textwidth
]
You are an expert multimodal reasoning model.

You are given an image and a corresponding text. Your task is to jointly analyze the visual and textual information to infer emotional expression, implicit intention, potential irony or sarcasm, and the deeper meaning conveyed by their interaction.

\medskip
\textbf{Input Text:}

\{\{ content | trim \}\}

\medskip
\textbf{Reasoning Guidelines:}

\begin{enumerate}
  \item \textbf{OCR Understanding.} Identify and interpret meaningful textual content appearing in the image, focusing on its emotional and semantic implications.
  \item \textbf{Visual Scene Analysis.} Examine visual cues such as emotional tone, atmosphere, attitude, social implications, and symbolic elements, taking into account exaggeration, contrast, metaphor, and contextual signals.
  \item \textbf{Cross-modal Integration.} Jointly reason over the external text, OCR text, and visual content to determine whether they are consistent, complementary, conflicting, or intentionally ironic.
  \item \textbf{High-level Reasoning.} Apply relevant background knowledge and commonsense reasoning to infer deeper communicative intent beyond surface-level interpretation.
\end{enumerate}

\medskip
\textbf{Output Specification.}

The model should output three fields:
\textit{ocr}, describing the OCR content with emotional and semantic interpretation;
\textit{visual\_scene}, examining visual cues such as emotional tone, atmosphere, attitude, social implications, symbolic elements, exaggeration, contrast, metaphor, and contextual signals;
and \textit{textimage\_relation}, explaining the relationship between the textual and visual content.
\end{tcolorbox}

\caption{Illustration of the multimodal reasoning prompt used in our experiments.}
\label{fig:multimodal_prompt}
\end{figure*}

\end{document}